\documentclass[letterpaper, 10 pt, conference]{ieeeconf}
\usepackage{amsmath,amsfonts}
\usepackage{algorithmic}
\usepackage{algorithm}
\usepackage{array}

\usepackage{textcomp}
\usepackage{stfloats}
\usepackage{url}
\usepackage{verbatim}
\usepackage{graphicx}
\usepackage{cite}
\usepackage{subcaption}
\usepackage{times}
\usepackage{booktabs}
\usepackage{multirow}
\usepackage{siunitx}
\usepackage[table]{xcolor}

\usepackage{multicol}
\usepackage[bookmarks=true]{hyperref}

\hypersetup{
    colorlinks=true,           
    urlcolor=blue,  
    citecolor=blue,
    linkcolor=black
}

\IEEEoverridecommandlockouts

\title{\LARGE \bf
120 Minutes and a Laptop: Minimalist Image-goal Navigation via Unsupervised Exploration and Offline RL
}

\author{Xiaoming Liu , Borong Zhang , Qingbiao Li , Steven Morad
\thanks{The authors are with Faculty of Science and Technology, University of Macau. Correspondence to: Steven Morad \textless smorad@um.edu.mo \textgreater 
}
\thanks{\textit{Preprint. March 27, 2026.}}
}
\begin{document}
\bstctlcite{IEEEexample:BSTcontrol}
\maketitle
\thispagestyle{empty}
\pagestyle{empty}

\begin{abstract}
The prevailing paradigm for image-goal visual navigation often assumes access to large-scale datasets, substantial pretraining, and significant computational resources. In this work, we challenge this assumption. We show that we can collect a dataset, train an in-domain policy, and deploy it to the real world (1) in less than 120 minutes, (2) on a consumer laptop, (3) without any human intervention. Our method, MINav, formulates image-goal navigation as an offline goal-conditioned reinforcement learning problem, combining unsupervised data collection with hindsight goal relabeling and offline policy learning. Experiments in simulation and the real world show that MINav improves exploration efficiency, outperforms zero-shot navigation baselines in target environments, and scales favorably with dataset size. These results suggest that effective real-world robotic learning can be achieved with high computational efficiency, lowering the barrier to rapid policy prototyping and deployment.
Video: \url{https://luckyxiaoming.github.io/MINav/}
\end{abstract}


\section{Introduction}
Image-goal visual navigation (ImageNav) is a fundamental capability for autonomous robots operating in real-world environments~\cite{zhu_targetdriven_2017}. Recently, this problem has seen substantial progress driven by large-scale visual navigation foundation models and Vision-Language-Action (VLA) models~\cite{shah_vint_2023,sridhar_nomad_2024, shah_lmnav_2022}. Trained on massive multi-embodiment datasets, these models have demonstrated impressive zero-shot generalization to novel environments and robotic platforms.

Despite these advances, deploying end-to-end ImageNav policies in a specific, unmapped physical environment still faces two key practical challenges: efficiency and platform specificity. Although recent foundation models and VLA models show promising cross-embodiment generalization, this capability does not directly translate into efficient deployment for ImageNav~\cite{shen_effonav_2025}. In practice, achieving reliable navigation performance in a specific real-world environment typically requires in-domain adaptation to the target robot and scene to bridge embodiment gaps and visual domain shifts~\cite{cui_grounded_2023}. However, such adaptation requires specialized trajectory formats and substantial computational resources, making rapid deployment difficult and expensive~\cite{hong_general_2024, shah_gnm_2023}. By contrast, learning a small specialized model purely from scratch in the target environment is an appealing alternative, but it is also challenging~\cite{mediratta_generalization_2023}. Its success depends on obtaining sufficiently diverse in-domain data under a limited real-world interaction budget. This raises a fundamental question: \textit{If target-domain data collection is ultimately unavoidable for a specific platform, can we achieve strong navigation performance through highly efficient, purely in-domain learning without relying on large-scale navigation pre-training or human demonstrations? }

\begin{figure}[!t]
    \centering
    \includegraphics[width=\linewidth]{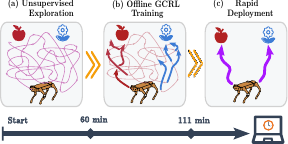}
    \caption{MINav enables rapid, two-hour deployment for real-world end-to-end ImageNav. (a) Autonomous data collection. (b) Offline policy learning. (c) Real-world policy deployment.}
    \label{figure:hero figure}
\end{figure}

In this paper, we show that robust end-to-end navigation can emerge from a lightweight policy trained on a minimal amount of autonomously collected real-world data. To this end, we formulate ImageNav as an offline goal-conditioned reinforcement learning (GCRL) problem. We address this challenge via two complementary principles: maximizing dataset coverage under a strict collection budget, and learning a robust policy entirely from real-world offline data, as shown in Figure~\ref{figure:hero figure}. Consequently, we propose Minimalist Image-goal Navigation (MINav), an efficient, fully automated pipeline for ImageNav. The main contributions of this work are summarized as follows: 
\begin{itemize}
\item We propose a method for real-world data collection, achieving broad dataset coverage without any human involvement.
\item We demonstrate the feasibility of offline GCRL for end-to-end ImageNav through real-world robotic experiments.
\item We integrate these contributions into MINav, a minimalist end-to-end pipeline that yields a deployable navigation policy on a laptop in under 120 minutes.
\item We show that MINav scales favorably as the amount of training data increases.
\item We demonstrate that MINav adapts to two different robot platforms without any modification.
\item We show that MINav remains robust in dynamic environments.
\end{itemize}

\begin{figure*}[t]
  \centering
\includegraphics[width=\linewidth]{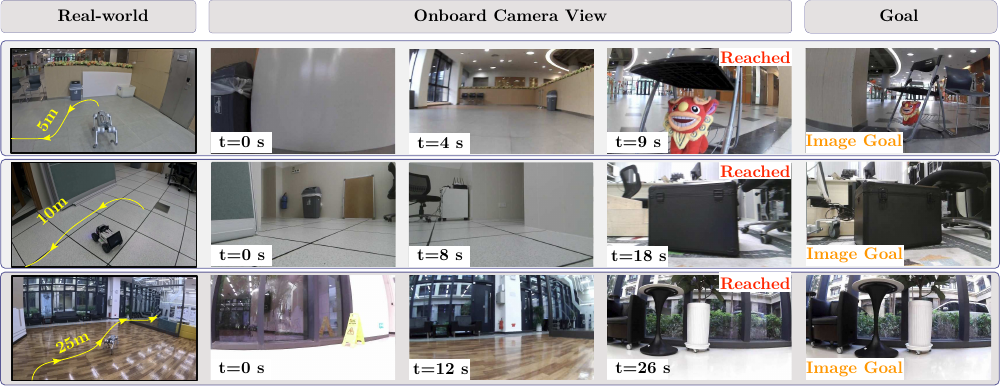} 
  \caption{Real-world deployment of MINav in the three evaluation environments. The robot successfully reaches image-goals despite clutter, blur, reflections, and partial occlusion.}
  \label{fig:experiments}
\end{figure*}

\section{Related Work}
\subsection{ImageNav Foundation Model}

Large-scale robot models and VLA models have recently been applied to robotic operations due to their strong generalization capabilities~\cite{intelligence_$p^_06$_2025, kim_openvla_2024}. Representative ImageNav foundation models include GNM, ViNT, and NoMaD. GNM supports zero-shot transfer across robots and operating conditions through a unified navigation model~\cite{shah_gnm_2023}; ViNT adopts an early-fusion transformer to encode observations and goal images jointly~\cite{shah_vint_2023}; and NoMaD extends ViNT with a masked diffusion objective in a transformer, achieving stronger performance~\cite{sridhar_nomad_2024}. Together, these works demonstrate the value of large-scale navigation pre-training for zero-shot transfer, though practical deployment in a specific target environment often still requires additional adaptation~\cite{hong_general_2024, shen_effonav_2025}. Moreover, adapting such large models can be computationally expensive and engineering-intensive, often requiring substantial GPU resources and additional action-space alignment~\cite{dettmers_qlora_2023, sridhar_nomad_2024}. In contrast, our work focuses on efficient deployment in a single unmapped real-world environment, where the goal is to rapidly acquire a robust policy from minimal in-domain data without any human intervention.

\subsection{Offline RL based ImageNav}
Offline reinforcement learning offers an appealing framework for robotic navigation because it enables policy learning directly from static datasets, without requiring additional online interaction~\cite{levine_offline_2020, morad_embodied_2021}. Prior work in offline RL has shown that goal-reaching behavior can be learned from relabeled trajectories~\cite{andrychowicz_hindsight_2017,shah_offline_2022}. However, most existing studies are developed and evaluated primarily on simulated datasets, where large-scale data collection is relatively cheap and controllable~\cite{zhou_real_2022, cui_grounded_2023}. Extending this paradigm to real-world ImageNav remains challenging. Training on broader real-world datasets may improve generalization~\cite{mediratta_generalization_2023}, but it also increases the cost of data collection and curation, while not necessarily yielding efficient deployment in a specific target environment due to potential domain mismatch~\cite{qiao_efficient_2026}. Our work instead studies a complementary setting, where the objective is not large-scale generalization, but fast deployment in a specific real-world environment.

\section{Preliminaries}
\subsection{Visual Navigation Task Formulation}
We consider a ImageNav task characterized by an observation space $\mathcal{O}$, a goal space $\mathcal{G}$, and a continuous action space $\mathcal{A}$. Both $\mathcal{O}$ and $\mathcal{G}$ consist of RGB images in $\mathbb{R}^{H \times W \times 3}$, where $H$ and $W$ denote the height and width, respectively. The objective is to learn a control policy $\pi$ that maps the current and historical observations $o_{t-k:t} \in \mathcal{O}$ and a target visual goal $g \in \mathcal{G}$ to an action $a_t \in \mathcal{A}$. This problem is formulated as a goal-conditioned Partially Observable Markov Decision Process (POMDP), defined by the tuple $\mathcal{P} = (\mathcal{S, O, A, G}, P, R, \gamma, H)$. Here, $\mathcal{S}$ represents the state space, $P(s_{t+1}|s_t, a_t)$ denotes the transition dynamics, and $R(s_t, a_t, g)$ is a goal-conditioned reward function. The term $\gamma \in [0, 1)$ is the discount factor. The policy $\pi$ aims to maximize the expected cumulative return:
\begin{equation}
J(\pi) = \mathbb{E}_{\tau \sim \pi} \left[ \sum_{t=0}^{\infty} \gamma^t R(s_t, a_t, g) \right],\end{equation}
where $\tau$ denotes the trajectory induced by the policy.

\subsection{Offline Reinforcement Learning}
Offline Reinforcement Learning (RL) aims to derive an optimal policy from a fixed dataset $\mathcal{D} = \{(s_t^{(i)}, a_t^{(i)}, s_{t+1}^{(i)}, r_t^{(i)})\}_{i=1}^N$ without active environment interaction~\cite{kumar_conservative_2020, eysenbach_contrastive_2022}. This dataset is pre-collected by an unknown behavior policy $\pi_\beta$. Because $\pi_\beta$ is often suboptimal or highly exploratory (e.g., a random policy), a distribution shift arises between the learning policy and the behavior policy. This shift frequently leads to overestimation of out-of-distribution actions. To mitigate this, algorithms such as TD3+BC regularize the learned policy to remain close to the original data distribution~\cite{fujimoto_minimalist_2021}. Specifically, the policy $\pi$ is optimized to jointly maximize the state-action value function $Q_\theta(s, a)$ while minimizing a behavioral cloning (BC) penalty:
\begin{equation}
J(\pi) = \mathbb{E}_{(s, a) \sim \mathcal{D}} \left[  Q_\theta(s, \pi(s)) - \lambda \| \pi(s) - a \|^2 \right],
\label{eq:td3bc_prelim}
\end{equation}

where $\lambda$ is a hyperparameter balancing reward maximization and BC.



\section{Minimalist Image-goal Navigation}
\begin{figure*}[!t]
  \centering
  \includegraphics[width=\linewidth]{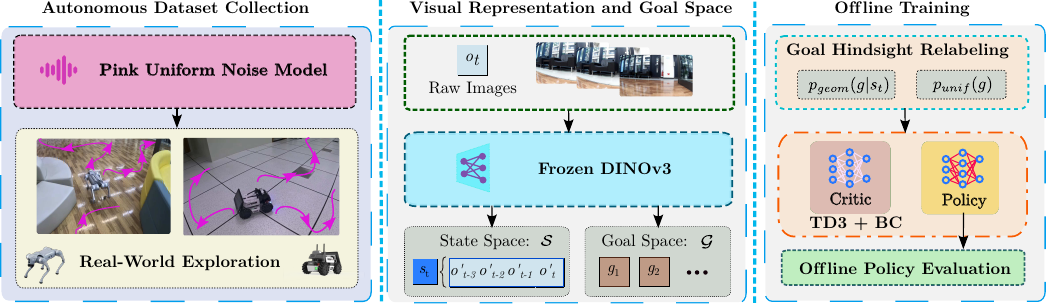} 
  \caption{Overview of the MINav pipeline. Starting from autonomous exploration in the target environment, MINav collects diverse trajectories using the proposed pink uniform noise model, extracts visual representations with frozen DINOv3, constructs the goal space, and trains a navigation policy offline via hindsight goal relabeling and TD3+BC.}
  \label{fig:overview}
\end{figure*}

We propose MINav, an efficient and fully automated pipeline for learning end-to-end in-domain ImageNav skills. As illustrated in Figure~\ref{fig:overview}, MINav makes no assumptions about the low-level controller, robot actuators, or camera specifics, enabling efficient data collection, curation, and policy training. The resulting policy can then be deployed in the same environment without any calibration. 

\subsection{Autonomous Dataset Collection via Pink Uniform Noise}
Since offline RL performance depends heavily on dataset coverage, we seek an exploration process that improves state-space expansion under limited real-world interaction budgets~\cite{mediratta_generalization_2023, hollenstein_action_2022}. To this end, we adopt pink noise as the base process for action generation, as its temporal correlation provides a better balance between randomness and consistency than white noise~\cite{eberhard_pink_2022}.

\begin{figure}[!t]
    \centering
    \includegraphics[width=\linewidth]{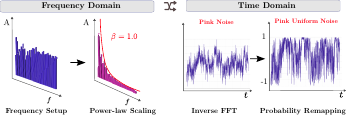} 
    \caption{Generation process of the proposed pink uniform noise.}
    \label{fig:generate noise}
\end{figure}

In practice, we generate a standard pink noise sequence by shaping a randomly initialized spectrum in the frequency domain to follow a power-law distribution with $\beta = 1.0$, and then transforming it back to the time domain via inverse FFT, as illustrated in Figure~\ref{fig:generate noise}.

Although the standard pink noise is effective for exploring the state space, it exhibits clear limitations in covering the action space. Since standard pink noise is derived from a gaussian distribution, the robot rarely samples actions near the boundaries of its control range (e.g., maximum velocity), which hinders the learning of optimal policies.

To address this issue, we apply the probability integral transform to obtain uniform marginal coverage over the action space while retaining the temporal dependence of the original pink noise. Specifically, if a random variable $Z \sim \mathcal{N}(0,1)$, then applying the cumulative distribution function (CDF) of the standard normal distribution, denoted by $\Phi$, yields $\Phi(Z) \sim \mathrm{Unif}(0,1)$. In our case, each sample of the pink-noise sequence is gaussian distributed, i.e., $x_t \sim \mathcal{N}(0,\sigma^2)$. Therefore, $\Phi(x_t/\sigma)$ is uniformly distributed on $[0,1]$. Since this transformation is deterministic and monotone, it preserves the temporal ordering of the original signal and thus retains its temporal dependence. We then linearly rescale the transformed signal to the action range: $a_t = a_{\min} + (a_{\max} - a_{\min}) \cdot \Phi\!\left(\frac{x_t}{\sigma}\right). $

 We refer to this exploration process as pink uniform noise. This design is particularly important in offline RL, where insufficient action-space coverage can lead to severe extrapolation error during policy optimization. More broadly, it is motivated by the convergence requirements of Q-learning, under which sufficiently extensive exploration is needed to obtain a near-optimal policy~\cite{kearns_finitesample_1998}.

\subsection{Visual Representation and Goal Space Construction}
\subsubsection{Visual Encoding via Frozen DINOv3}
Learning a visual encoder is computationally expensive. To mitigate this issue, we adopt DINOv3 as a frozen visual backbone~\cite{simeoni_dinov3_2025}. DINOv3 is pre-trained via self-distillation, capturing semantic geometry and structural correspondences robust to viewpoint changes. Let $\phi$ denote the normalized feature extraction function induced by DINOv3. The visual representation of an image observation $o_t$ is defined as $o'_t = \phi(o_t)$.

\subsubsection{Valid Goal Selection via Spatial Standard Deviation}

Since the exploratory policy is random, the collected dataset $\mathcal{D}$ lacks explicit navigational goals. While future observations can be reused as goals, not all such observations are valid goal candidates. In particular, observations captured when the robot is extremely close to obstacles, such as a wall, often contain insufficient structural information to serve as meaningful goals~\cite{krantz_instancespecific_2022}. We similarly observe that selecting such observations as goals degrades agent performance. 

\begin{figure}[!t]
    \centering
    \includegraphics[width=\linewidth]{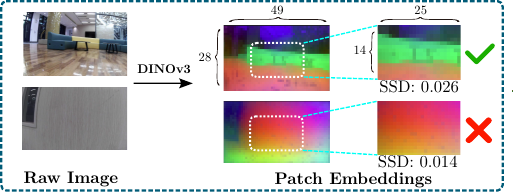} 
    \caption{Valid goal selection using DINOv3-based spatial standard deviation.}
    \label{fig:goal filter}
\end{figure}

To remove uninformative goals, we define a DINOv3-based spatial standard deviation (SSD) metric. Given a $448 \times 784$ input image, DINOv3 produces a $28 \times 49$ grid of patch embeddings from its last hidden layer, as shown in Figure~\ref{fig:goal filter}. We compute the SSD, denoted as $\sigma_{spa}(o)$, over the embeddings in the center-cropped region. A low SSD indicates a visually homogeneous observation with limited structural content, which is therefore excluded from the valid goal space.

For the entire dataset, we construct a global valid goal set $\mathcal{G}$ by only retaining observations whose SSD exceeds a predefined spatial threshold $\delta_{SSD}$ of 0.02. Formally, this filtered goal set  $\mathcal{G}$ is defined as $ \mathcal{G} = \{ \phi(o) \mid o \in \mathcal{D}, \, \sigma_{{spa}}(o) > \delta_{SSD} \} $.

\subsection{State, Goal, and Reward Design}
To learn navigation skills from the static dataset, we formulate ImageNav as an offline GCRL problem over the extracted visual representation space, with carefully designed state, goal and reward definitions.

\subsubsection{State Representation}
To approximate the Markov property under partial observability, we define the RL state $s_t$ as the concatenation of visual representation from four consecutive observations $s_t = [o'_{t-3}, o'_{t-2}, o'_{t-1}, o'_t] $.
Stacking these historical frames provides the policy with essential temporal context, implicitly capturing the robot's velocity and acceleration information.

\subsubsection{Hindsight Goal Relabeling}
To provide dense and meaningful supervision, we employ a hindsight relabeling strategy. For a given transition $(s_t, a_t, s_{t+1})$ sampled from a trajectory of length $T$ within the dataset $\mathcal{D}$,  we dynamically sample a hindsight goal $g$ using a mixture of two state distributions:
\begin{itemize}
\item  \emph{Geometric Future Sampling} ($p_{geom}(g|s_t)$):  given the current timestep $t$, we sample an offset $k$ from a geometric distribution $P(K=k) = p^{k-1}(1-p)$ for $k \ge 1$, and set the hindsight goal to $g = o'_{\min(t+k, T)}$.
\item \emph{Global Uniform Sampling} ($p_{unif}(g)$): we uniformly sample a valid goal from the global goal set $\mathcal{H}$. 
\end{itemize}

In practice, we construct a mixture distribution $p_{mix}(g|s_t)$ by combining these two strategies. Specifically, we use a ($0.5, 0.5$) mixture for critic training and uniform sampling only for actor training.

\subsubsection{Goal-Conditioned Reward}
We evaluate the success of the task by calculating the average cosine similarity between recent visual representations in the stacked state $s_t$ and the goal $g$. Averaging the similarities over four consecutive visual representations provides behavioral smoothing. The state-goal similarity $S(s_t, g)$ is defined as: 
\begin{equation}
    S(s_t, g) = \frac{1}{4} \sum_{k=t-3}^{t} \text{sim}_{\text{cos}}(o'_k, g)
\end{equation}
Once $S(s_t, g)$ exceeds the predefined threshold $\delta_{\text{done}}$, the agent receives a sparse reward $R(s_t, g)=1$ if $S(s_t, g)\ge \delta_{\text{done}}$, and $R(s_t, g)=0$ otherwise, where $\delta_{\text{done}}=0.8$; upon receiving the reward, the terminal flag is set to $d=1$.

\subsection{Offline Policy Learning and Model Selection}
\subsubsection{Policy Optimization via TD3+BC}
Given the relabeled goal-conditioned transitions and the sparse reward function, we optimize our navigation policy using a minimalist offline RL algorithm based on TD3+BC. Specifically, the critic networks $Q_{\theta_i}(s_t, a_t, g)$ for $i \in \{1, 2\}$ are trained to estimate the expected cumulative return by minimizing the expected temporal difference (TD) objective:
\begin{equation}
    J(\theta_i) = \mathbb{E}_{\substack{(s_t, a_t, s_{t+1}) \sim \mathcal{D} \\ g \sim p_{\text{mix}}(g|s_t)}} \left[ \left( Q_{\theta_i}(s_t, a_t, g) - y_t \right)^2 \right],
\end{equation}
where the target value $y_t$ is computed using clipped double Q-learning using the slowly updated target networks:
\begin{equation}
    y_t = R(s_t, g) + \gamma \min_{j=1,2} Q_{\theta'_j}(s_{t+1}, \pi_{\phi'}(s_{t+1}, g) + \epsilon, g).
\end{equation}
Subsequently, by adapting Eq.~\ref{eq:td3bc_prelim} to our goal-conditioned setting, the actor objective $J(\pi_\phi)$ is to maximize:
\begin{equation}
    J(\pi_\phi) = \mathbb{E}_{\substack{(s_t, a_t) \sim \mathcal{D} \\ g \sim p_{\text{unif}}(g)}} \left[ Q_{\theta_1}(s_t, \hat{a}_t, g) - \lambda \| \hat{a}_t - a_t \|^2 \right]
\end{equation}
where $\hat{a}_t = \pi_\phi(s_t, g)$ is the action conditioned on the goal predicted by the policy, and $\lambda$ is set at 0.001.

\subsubsection{Offline Model Selection via Fitted Q-Evaluation}
Since the policy is trained fully offline, it is critical to identify the best-performing policy checkpoint without requiring further interactions, as the real-world evaluation is expensive. Effective checkpoint selection allows us to stop training at the appropriate time and deploy the robot rapidly. To achieve this, we employ Fitted Q-Evaluation (FQE) as an offline validation metric. 

Given a saved policy checkpoint $\pi_\phi$, FQE trains an independent evaluation Q-network, denoted $Q_\omega^{\pi_\phi}$, to estimate the expected return $Q_\omega^{\pi_\phi}(s_0, a_0, g)$ of executing the policy from an initial state-action-goal tuple. This is achieved by minimizing the TD error:
\begin{equation}
    J(\omega) = \mathbb{E}_{\substack{(s_t,a_t, s_{t+1}) \sim \mathcal{D} \\ g \sim p_{\text{unif}}(g)}} \left[ \left( Q_\omega^{\pi_\phi}(s_t, a_t, g) - y^{\text{FQE}}_t \right)^2 \right],
\end{equation}
where the evaluation target is defined as:
\begin{equation}
    y^{\text{FQE}}_t = R(s_t, g) + \gamma Q_{\omega'}^{\pi_\phi}(s_{t+1}, \pi_\phi(s_{t+1}, g), g).
\end{equation}
Unlike the training critic, the FQE target depends exclusively on the action proposed by the evaluated policy, rather than the behavior action stored in the dataset. We use the estimated policy value under FQE to rank checkpoints and select the final model for deployment.

 \begin{figure*}[htbp]
  \centering
  \includegraphics[width=0.98\textwidth]{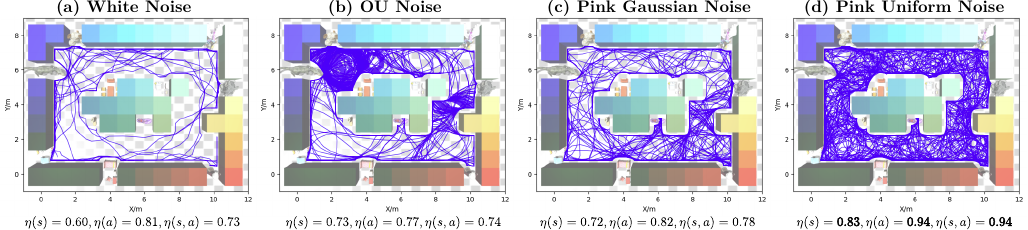} 
  \caption{Visualization of exploration trajectories generated by different noise strategies. The corresponding normalized state entropy $\eta(s)$, action entropy $\eta(a)$, and state-action joint entropy $\eta(s, a)$ are reported at the bottom of each subfigure, highlighting the superior coverage of the pink uniform strategy.}
  \label{fig:exploration}
\end{figure*}

\section{Experiment}
We evaluate MINav in both MuJoCo simulation and real-world environments to answer the following key questions:
\begin{itemize}
    \item \textbf{Q1:} Does the proposed unsupervised exploration strategy achieve more efficient exploration than standard noise baselines under a fixed interaction budget?
    \item \textbf{Q2:} How quickly can MINav produce a usable policy?
    \item \textbf{Q3:} How does MINav performance scale with dataset size?
    \item \textbf{Q4:} Does MINav remain robust across different robot platforms and in the presence of dynamic human interference?

\end{itemize}

\subsection{Experimental Setup}
We consider a mapless ImageNav setting in which the policy observes only RGB images from an onboard camera. In real-world experiments, LiDAR is used solely for collision prevention, namely to stop forward motion when an obstacle is detected within a predefined threshold. LiDAR is never included in the policy observation space.
The policy control the robot through a high-level velocity interface. The continuous 3-dimensional action space is defined as $a_t = [v_x, v_y, \omega_z] \in [-1, 1]^3$, where $v_x$ is the surge velocity mapped to $[0, v_{x_{\max}}]$ m/s, $v_y$ is the sway velocity mapped to $[-v_{y_{\max}}, v_{y_{\max}}]$ m/s, and $\omega_z$ is the yaw angular velocity mapped to $[-\omega_{\max}, \omega_{\max}]$ rad/s. 

\subsection{ Exploration Efficiency and Dataset Quality} 
To answer \textbf{Q1} and \textbf{Q3}, we first evaluate the proposed exploration strategy in MuJoCo simulation. Our simulated environment is built upon the PointMaze benchmark, with modifications to the observation space, action space, and goal to better align with the real-world ImageNav setting~\cite{park_ogbench_2025, downs_google_2022}.  We compare pink uniform noise against pink Gaussian noise, Ornstein-Uhlenbeck (OU) noise~\cite{uhlenbeck_theory_1930}, and white noise. All datasets are compared under the same interaction budgets. We evaluate each strategy from two perspectives: dataset coverage and downstream policy performance.

To quantitatively evaluate the dataset coverage, we compute the normalized state entropy $\eta(s)$, normalized action entropy $\eta(a)$, and normalized state-action joint entropy $\eta(s, a)$. In simulation, we have access to the ground-truth state. Therefore, we use the agent’s spatial coordinates as the ground-truth state when computing these entropy metrics. We define normalized entropy $\eta(z)$ as:
\begin{equation}
    \eta(z) = \frac{-\sum_{j=1}^{K} p(\hat{z}_j) \log_2 p(\hat{z}_j)}{\min(\log_2 K, \log_2 N)},
\end{equation}
where $\hat{z}$ denotes the discretized representation with an adaptive resolution $\Delta \propto N^{-1/P}$, and $K$ denotes the total number of bins. Here, $N$ is the total dataset size and $P$ is the dimensionality of $z$. 

As shown in Figure~\ref{fig:exploration}, white noise strategy tends to get trapped near the boundaries of the environment, leading to poor coverage of the central area. OU noise exhibits strong local correlation but exhibits relatively low action entropy. Both pink gaussian noise and pink uniform noise cover the overall state space more effectively, but they differ substantially in trajectory density. In particular, the pink uniform noise strategy produces both better action and state coverage, leading to more complete trajectory coverage. Accordingly, it consistently attains the highest normalized state entropy, action entropy, and state–action joint entropy among all baselines.
\begin{figure}[!t]
  \centering
\includegraphics[width=0.48\textwidth]{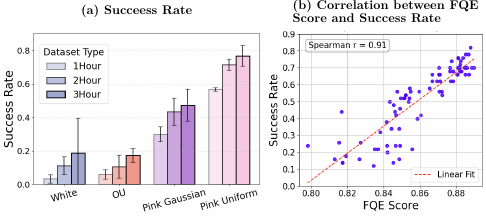} 
  \caption{Simulation evaluation of downstream performance and offline model selection. (a) Success rates of offline policies trained on datasets collected with different exploration strategies. (b) Correlation between FQE scores and ground-truth success rates, showing that FQE is a reliable proxy for offline checkpoint selection.}
  \label{fig:SR and FQE}
\end{figure}

We next evaluate the downstream navigation performance of policies trained on these datasets. Using the same training pipeline across all four datasets and three random seeds, we train identical agents and evaluate each policy on 10 distinct simulation goals, with 5 randomized starting positions per goal.
As shown in Figure~\ref{fig:SR and FQE} (a), policies trained on pink uniform noise consistently outperform those trained on the baseline datasets across all data budgets. Notably, an agent trained on just 1 hour of pink uniform  data  achieves a success rate comparable to one trained on 3 hours of pink gaussian data. This highlights the superior exploration efficiency of our method. Furthermore, MINav exhibits a clear positive scaling trend: as the dataset budget increases from 1 to 3 hours, the downstream success rate improves steadily. 

To validate our offline model selection strategy, we further analyze the correlation between FQE scores and ground-truth success rates. We extract policy checkpoints at regular intervals during training from the pink uniform experiments under 3-hour budgets. For each checkpoint, we compute its FQE score and measure its actual simulation success rate. As visualized in Figure.~\ref{fig:SR and FQE} (b), the FQE score exhibits a strong linear correlation with the performance of the downstream policy. Fitting the relationship and computing the Spearman’s rank correlation yields a coefficient of 0.91, indicating that FQE is a reliable proxy for the true policy success rate and can support model selection without additional online interaction.

\subsection{Real-World Visual Navigation Experiments} 

For \textbf{Q2} and \textbf{Q3}, we evaluate MINav on the Unitree Go2 quadruped robot in real-world indoor environments. Following the simulation setup, we apply the pink uniform noise strategy to the robot’s velocity commands during autonomous data collection. Although the robot’s low-level controller runs at a higher frequency, the exploration policy is executed at \SI{2}{\hertz}. A velocity smoothing filter is used to upsample commands to \SI{20}{\hertz}, ensuring stable and jitter-free locomotion.

\subsubsection{Experimental Scenarios and Data Collection}
\begin{figure}[!t]
  \centering
\includegraphics[width=0.48\textwidth]{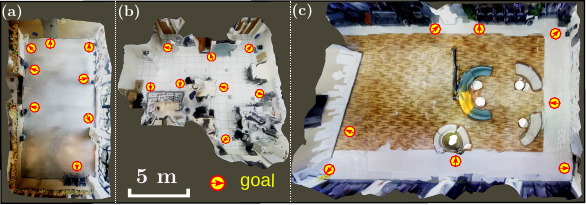} 
  \caption{Real-world evaluation environments. We consider three indoor environments of increasing difficulty: (a) Simple, (b) Standard, and (c) Complex.}
  \label{fig:map}
\end{figure}

As shown in Figure~\ref{fig:map}, we designed three distinct indoor environments with increasing difficulty:
\begin{itemize}
\item \emph{Scenario A }(Simple): A \SI{60}{\square\meter} empty room.
\item \emph{Scenario B }(Standard): A \SI{60}{\square\meter} office with dense furniture.
\item \emph{Scenario C }(Complex): A \SI{200}{\square\meter} room with more complex obstacles and structural variation.
\end{itemize}

For each environment, we collected a specific 2-hour dataset consisting only of RGB observations and executed velocity commands. To fully exploit the DINOv3 visual transformer, raw RGB frames are recorded at $720 \times 1280$ resolution and then resized to $448 \times 784$ for feature extraction.

The training process was conducted on a consumer laptop (NVIDIA RTX 5070 GPU, 32~GB RAM). We used a batch size of 1024 and trained the policy for 50,000 gradient steps, computing FQE every 1,000 steps.  We then utilized the FQE metric to reliably select the optimal policy checkpoint for deployment. For real-world deployment, we defined 8 spatially diverse goals per environment. For each goal, we conducted 5 trials with randomized starting positions, setting the maximum episode length to 40–80 steps depending on the room difficulty.

\begin{table*}[t]

\centering

\caption{Real-world ImageNav performance across three environments. }

\label{tab:overal performance}

\resizebox{\textwidth}{!}{

\begin{tabular}{llccc ccc ccc ccc} 

\toprule

\multirow{2}{*}{\textbf{Method}} & \multirow{2}{*}{\textbf{Visual backbones}} & \multirow{2}{*}{\shortstack{\textbf{Policy} \\ \textbf{Params} $\downarrow$}} & \multirow{2}{*}{\shortstack{\textbf{Training Cost} \\ \textbf{(TFLOPs-hours)} $\downarrow$}} & \multicolumn{3}{c}{\textbf{Simple}} & \multicolumn{3}{c}{\textbf{Standard}} & \multicolumn{3}{c}{\textbf{Complex}} \\ 

\cmidrule(lr){5-7} \cmidrule(lr){8-10} \cmidrule(lr){11-13} 

& & & & \textbf{SR($\%$)} $\uparrow$ & \textbf{Time($s$)} $\downarrow$ & \textbf{STL($\%$)} $\uparrow$ & \textbf{SR($\%$)} $\uparrow$ & \textbf{Time($s$)} $\downarrow$ & \textbf{STL($\%$)} $\uparrow$ & \textbf{SR($\%$)} $\uparrow$ & \textbf{Time($s$)} $\downarrow$ & \textbf{STL($\%$)} $\uparrow$ \\ 

\midrule

Random         & N/A         & N/A    & N/A    & $0\: \scriptstyle{\pm 0}$  & $20\: \scriptstyle{\pm 0}$  & $0\: \scriptstyle{\pm 0}$  & $0\: \scriptstyle{\pm 0}$ & $30\: \scriptstyle{\pm 0}$ & $0\: \scriptstyle{\pm 0}$ & $0\: \scriptstyle{\pm 0}$  & $40\: \scriptstyle{\pm 0}$  & $0\: \scriptstyle{\pm 0}$  \\

NoMad          & EfficientNet   & $19\: M$ & $>30000$   & $18\: \scriptstyle{\pm 7}$ & $18\: \scriptstyle{\pm 1}$ & $14\: \scriptstyle{\pm 7}$ & $18\: \scriptstyle{\pm 19}$ & $25\: \scriptstyle{\pm 4}$ & $16\: \scriptstyle{\pm 18}$ & $18\: \scriptstyle{\pm 11}$ & $37\: \scriptstyle{\pm 3}$ & $16\: \scriptstyle{\pm 10}$ \\

ViNT           & EfficientNet& $30\: M$ & $30000$   &$8\: \scriptstyle{\pm 7}$ & $6\: \scriptstyle{\pm 5}$ & $19\: \scriptstyle{\pm 1}$ & $13\: \scriptstyle{\pm 9}$ & $28\: \scriptstyle{\pm 3}$ & $9\: \scriptstyle{\pm 7}$ & $8\: \scriptstyle{\pm 17}$ & $38\: \scriptstyle{\pm 4}$ & $7\: \scriptstyle{\pm 15}$ \\

GNM            & MobileNetV2 & $9\: M$  & $>5000$   & $9\: \scriptstyle{\pm 12}$  & $18\: \scriptstyle{\pm 1}$  & $9\: \scriptstyle{\pm 11}$  & $10\: \scriptstyle{\pm 10}$ & $28\: \scriptstyle{\pm 2}$ & $7\: \scriptstyle{\pm 10}$ & $25\: \scriptstyle{\pm 13}$ & $36\: \scriptstyle{\pm 3}$ & $22\: \scriptstyle{\pm 11}$ \\

\rowcolor{blue!60!purple!5} Ours (Minimal, 1h) & DINOv3-ViT-S & $\textbf{2}\: M$ & $150$ & $73\: \scriptstyle{\pm 10}$ & $12\: \scriptstyle{\pm 3}$ & $60\: \scriptstyle{\pm 13}$ & $43\: \scriptstyle{\pm 2}$ & $24\: \scriptstyle{\pm 2}$ & $29\: \scriptstyle{\pm 11}$ & $25\: \scriptstyle{\pm 9}$ & $33\: \scriptstyle{\pm 9}$ & $21\: \scriptstyle{\pm 7}$ \\ 

\rowcolor{blue!80!purple!10} Ours (Minimal, 1h) & DINOv3-ViT-L & $\textbf{2}\: M$ & $150$ & $85\: \scriptstyle{\pm 6}$ & $10\: \scriptstyle{\pm 1}$ & $70\: \scriptstyle{\pm 4}$ & $38\: \scriptstyle{\pm 15}$ & $26\: \scriptstyle{\pm 6}$ & $30\: \scriptstyle{\pm 12}$ & $38\: \scriptstyle{\pm 15}$  & $31\: \scriptstyle{\pm 2}$  & $30\: \scriptstyle{\pm 10}$  \\ 

\rowcolor{blue!60!purple!5} Ours (Extended, 2h) & DINOv3-ViT-S & \textbf{2} M & $150$ & $\textbf{95}\: \scriptstyle{\pm 7}$ & $8\: \scriptstyle{\pm 1}$ & $85\: \scriptstyle{\pm 7}$ & $\textbf{88}\: \scriptstyle{\pm 9}$ & $\textbf{16}\: \scriptstyle{\pm 3}$ & $\textbf{66}\: \scriptstyle{\pm 11}$ & $58\: \scriptstyle{\pm 14}$ & $27\: \scriptstyle{\pm 3}$ & $48\: \scriptstyle{\pm 12}$ \\ 

\rowcolor{blue!80!purple!10} Ours (Extended, 2h) & DINOv3-ViT-L & $\textbf{2}\: M$ & $150$ & $\textbf{100}\: \scriptstyle{\pm 0}$ & $\textbf{7}\: \scriptstyle{\pm 3}$ & $\textbf{92}\: \scriptstyle{\pm 11}$ & $\textbf{85}\: \scriptstyle{\pm 6}$ & $\textbf{15}\: \scriptstyle{\pm 3}$ & $\textbf{68}\: \scriptstyle{\pm 11}$ & $\textbf{85}\: \scriptstyle{\pm 10}$ & $\textbf{21}\: \scriptstyle{\pm 4}$  & $\textbf{66}\: \scriptstyle{\pm 10}$  \\ 

\bottomrule
\vspace{0.25em}
\end{tabular}
}
\captionsetup{font=small}
\caption*{For each environment, results are first averaged over 8 goals in each trial round, then reported as mean $\pm$ standard deviation over 5 rounds with different starting points. We compare MINav against zero-shot visual navigation baselines under varying data budgets and visual backbones. Values within 95\% of the best result in each column are shown in bold.}
\end{table*}

\subsubsection{Baselines}
We compare our method against three state-of-the-art visual navigation foundation models: \textbf{GNM}, \textbf{ViNT}, and \textbf{NoMaD}. These models are deployed in a zero-shot manner on our quadruped platform. This comparison allows us to benchmark large generalist models against our fast in-domain learning approach.

\subsubsection{Evaluation Metrics}
During evaluation, we use the state-goal similarity as an automatic proxy for goal attainment, and classify an episode as successful when the similarity exceeds 0.75. The threshold was determined through manual calibration to best match true goal-reaching. We further manually reviewed all evaluation outcomes to confirm that the episodes counted as successful indeed correspond to true arrivals. Therefore, the reported success rate is not solely based on the representation-space metric, but is also validated by human judgment.

Because accurate path-length measurements are unavailable in our setup, the standard SPL metric cannot be applied. Instead, we adopt Success Rate (SR), Completion Time, and Success weighted by Time Length (STL) as our primary proxy metrics. STL jointly evaluates navigation success and efficiency, defined as:
\begin{equation}
STL = \frac{1}{N}\sum_{i=1}^{N} S_i \frac{T_i^{*}}{\max(T_i, T_i^{*})} 
\end{equation}
where $N$ is total number of evaluation episodes and $S_i \in \{0,1\}$ indicates whether episode $i$ is successful. $T_i$ represents the robot's completion time, while $T_i^*$ denotes the reference completion time for the corresponding goal, obtained by averaging five human-expert trials for the same goal.

\begin{table}[!t]
\centering
\caption{End-to-end runtime of MINav. The full pipeline is completed in under 120 minutes on a laptop.}
\label{tab:time_cost}
\resizebox{\linewidth}{!}{
\begin{tabular}{cccc}
\toprule
 \textbf{Data Collection} & \textbf{Data Processing} & \textbf{Policy Training} & \textbf{Total Time} \\
\midrule
  60 min & 3 min (ViT-S) & 22 min & \textbf{85 min} \\
 60 min & 22 min (ViT-L) & 29 min & \textbf{111 min} \\
 120 min & 6 min (ViT-S) & 22 min & \textbf{148 min} \\
 120 min & 44 min (ViT-L) & 29 min & \textbf{193 min} \\
\bottomrule
\end{tabular}
}
\end{table}

\subsubsection{Experiment Results} 
We distinguish between a minimal deployment setting (1-hour collection, 111-minute full pipeline) and an extended data setting (2-hour collection) used to study scaling and report the strongest performance. Table~\ref{tab:overal performance} summarizes the quantitative results across all environments. Although the zero-shot baselines are trained on massive datasets, they struggle to achieve reliable navigation in the target environments. Among the baselines, GNM achieves the highest SR among the baselines, yet it only reaches 25\% in the complex scenario. In contrast, MINav trained on only 1 hour of in-domain data already outperforms all baselines across nearly all metrics and environments, as shown in Fugure~\ref{fig:experiments}. For comparison, GNM is trained on a dataset containing 70 hours of robot data and achieves an SR of 9\%, whereas our ViT-S-based variant, trained with only 1 hour of in-domain data, achieves an SR of 73\% in the simple scenario. We emphasize that this comparison is deployment-oriented: MINav uses a small amount of target-environment data, whereas the foundation-model baselines are evaluated zero-shot. The goal is therefore to compare practical deployment efficiency in a specific environment rather than adaptation-matched generalization. Thus, these results suggest that, for practical deployment in a specific environment, rapid in-domain learning with a small but high-quality dataset can be more effective than relying solely on zero-shot generalization from large foundation models.

The results also highlight a strong positive scaling trend with respect to the dataset size. When the training budget increases from 1-hour to 2-hour, performance improves substantially. Notably, in the Standard environment, the SR of our ViT-S based variant jumps from 43\% to 88\%, and its STL improves from 29\% to 66\%. This confirms that MINav benefits directly from additional autonomous data collection.

We also analyze the effect of encoder scale by comparing distilled DINOv3 ViT-S and ViT-L backbones. Interestingly, the lightweight ViT-S performs comparably to the larger ViT-L in most scenarios, with a noticeable gap emerging only in the complex environment under the 2-hour setting. Given the substantially lower inference cost of ViT-S, this result suggests that MINav paired with a compact visual encoder is a strong practical choice for resource-constrained robotic platforms.

Notably, Table~\ref{tab:time_cost} reports the total end-to-end time cost of the pipeline on the laptop. From 1-hour data collection to deployment, the full process can be completed in under two hours on a laptop, further supporting the practicality of the proposed minimalist framework.

\subsubsection{Impact of Noise Distribution}
Although the results above show that MINav benefits from increased data, the quality of the collected data is equally important. To isolate the effect of exploration during data collection, we conduct a real-world ablation in which three 1-hour datasets are collected using white uniform noise, pink Gaussian noise, and pink uniform noise, respectively, while keeping the downstream learning pipeline unchanged.

Similar to the simulation experiments (Figure~\ref{fig:SR and FQE}), the choice of exploration noise has a clear impact on real-world downstream policy performance under a fixed collection budget (Table~\ref{tab:noise_comparison}). Policies trained on data collected with white uniform noise perform poorly, achieving only 15\% SR, largely because the resulting trajectories fail to cover the state space effectively within one hour. Pink Gaussian noise performs better, but still underperforms pink uniform noise. Among the three strategies, pink uniform noise achieves the best overall results, with 43\% SR and 29\% STL. Compared with pink Gaussian noise, the uniform variant provides broader and more balanced coverage over the action range, reducing over-concentration on low-velocity actions. These results indicate that, under limited real-world data budgets, combining temporal correlation with bounded action-space coverage produces more useful datasets for downstream offline policy learning.

\begin{table}[!t]
\centering
\caption{Real-world ablation on exploration noise under a fixed 1-hour data collection budget.}
\label{tab:noise_comparison}
\resizebox{\linewidth}{!}{
\begin{tabular}{c ccc}
\toprule
\textbf{Metric} &\textbf{SR ($\%$}) $\uparrow$  & \textbf{Time ($s$)} $\downarrow$  & \textbf{STL ($\%$)} $\uparrow$  \\
\midrule
White Uniform Noise   &  $15\: \scriptstyle{\pm 10}$ &  $26\: \scriptstyle{\pm 3}$ &  $14\: \scriptstyle{\pm 10}$ \\
Pink Gaussian Noise &  $30\: \scriptstyle{\pm 14}$ &  $24\: \scriptstyle{\pm 3}$ &  $27\: \scriptstyle{\pm 14}$ \\
\cellcolor{blue!80!purple!15}Pink Uniform Noise &\cellcolor{blue!80!purple!15}$43\: \scriptstyle{\pm 2}$ & \cellcolor{blue!80!purple!15}$24\: \scriptstyle{\pm 2}$ & \cellcolor{blue!80!purple!15}$29\: \scriptstyle{\pm 11}$ \\
\bottomrule
\end{tabular}
}
\end{table}

\subsection{Robustness in Dynamic Environments Across Platforms}
To answer \textbf{Q4}, we evaluate MINav under dynamic human interference and on a different robotic platform (AgileX Limo wheeled robot). For both the quadruped and wheeled robots, we collect a 2-hour dataset in the Standard environment. To isolate the effect of dynamic interference, we evaluate only a subset of five goals that were highly reliable under static conditions. During dynamic trials, a pedestrian continuously walks, both obstructing the robot's path and remaining within camera view for the duration of the task, introducing persistent visual and spatial disturbances.

As reported in Table~\ref{tab:generalization_rearranged}, MINav achieves strong performance in both static and dynamic conditions, demonstrating robustness to human motion and effective transfer across embodiments. For the Go2 quadruped, dynamic interference slightly reduces efficiency while preserving a 100\% SR. For the Limo wheeled robot, all metrics decrease modestly under dynamic interference, yet the policy remains reliable overall. These results show that MINav is robust to hardware-specific variations and real-world visual perturbations.

\begin{table}[t]
\centering
\caption{Robustness of MINav across robot platforms and environmental conditions.}
\label{tab:generalization_rearranged}
\resizebox{\linewidth}{!}{
\begin{tabular}{ll ccc}

\toprule
\textbf{Platform} & \textbf{Condition} & \textbf{SR($\%$)} $\uparrow$ & \textbf{Time($s$)} $\downarrow$ & \textbf{STL($\%$)} $\uparrow$ \\
\midrule
\multirow{2}{*}{\shortstack{Quadruped \\ (Go2)}} 
    & Static  &  $100\: \scriptstyle{\pm 0}$ & $13\: \scriptstyle{\pm 4}$ & $92\: \scriptstyle{\pm 8}$ \\
    \cmidrule{2-5}
    & Dynamic & \cellcolor{blue!80!purple!10}$100\: \scriptstyle{\pm 0}$ & \cellcolor{blue!80!purple!10}$14\: \scriptstyle{\pm 2}$ & \cellcolor{blue!80!purple!10}$91\: \scriptstyle{\pm 3}$ \\

\midrule
\multirow{2}{*}{\shortstack{Wheeled \\ (Limo)}}    
    & Static  &  $92\: \scriptstyle{\pm 18}$ &  $8\: \scriptstyle{\pm 3}$ &  $91\: \scriptstyle{\pm 18}$ \\
    \cmidrule{2-5}
    & Dynamic & $88\: \scriptstyle{\pm 11}$ &  $14\: \scriptstyle{\pm 3}$ &  $83\: \scriptstyle{\pm 12}$ \\

\bottomrule
\end{tabular}
}
\end{table}

\section{Conclusion}
We presented MINav, a minimalist and fully automated framework for real-world image-goal navigation. By combining efficient autonomous exploration, frozen visual representations, and offline GCRL, MINav learns robust end-to-end navigation policies from a small amount of in-domain data without any human intervention. Experiments in simulation and the real world show that MINav achieves strong exploration efficiency, scales with increasing data, outperforms zero-shot baselines in target environments, and remains robust across robot platforms and dynamic human interference. These results suggest that efficient in-domain offline learning is a practical and effective alternative for specific robotic deployments.


\bibliographystyle{IEEEtran}
\bibliography{references}



\end{document}